\documentclass[conference]{IEEEtran}

\usepackage{amsmath,amsfonts}
\usepackage{algorithmic}
\usepackage{algorithm}
\usepackage{array}
\usepackage[caption=false,font=normalsize,labelfont=sf,textfont=sf]{subfig}
\usepackage{textcomp}
\usepackage{stfloats}
\usepackage{url}
\usepackage{verbatim}
\usepackage{graphicx}
\usepackage{cite}

\usepackage{amssymb}
\usepackage{mathtools}
\usepackage{adjustbox}
\usepackage{multirow} 
\usepackage{xurl} 

\usepackage{subfig}
\usepackage{pgf,tikz,pgfplots}
\pgfplotsset{compat=1.15}
\usepackage{mathrsfs}
\usetikzlibrary{arrows}
\usepackage{tikz-3dplot} 
\usetikzlibrary{matrix,decorations.pathreplacing} 

\usepackage{booktabs} 

\usepackage{enumitem}

\usepackage[colorlinks=true,linkcolor=blue]{hyperref}
\hypersetup{
    colorlinks=true,
    citecolor=blue,
    linkcolor=blue,
    filecolor=blue,      
    urlcolor=blue
    }

\begin{document}
    
    \title{Generating Topologically and Geometrically Diverse Manifold Data in Dimensions Four and Below}
    
    \author{\IEEEauthorblockN{Khalil Mathieu Hannouch}
        \IEEEauthorblockA{\textit{School of Information and Physical Sciences} \\
        \textit{The University of Newcastle, University Drive}\\
        Callaghan, NSW 2308, Australia \\
        khalil.hannouch@uon.edu.au}
        \thanks{Copyright 2024 IEEE. Published in the Digital Image Computing: Techniques and Applications, 2024 (DICTA 2024), 27-29 November 2024 in Perth, Western Australia, Australia. Personal use of this material is permitted. However, permission to reprint/republish this material for advertising or promotional purposes or for creating new collective works for resale or redistribution to servers or lists, or to reuse any copyrighted component of this work in other works, must be obtained from the IEEE. Contact: Manager, Copyrights and Permissions / IEEE Service Center / 445 Hoes Lane / P.O. Box 1331 / Piscataway, NJ 08855-1331, USA. Telephone: + Intl. 908-562-3966.}
        \and
        \IEEEauthorblockN{Stephan Chalup}
        \IEEEauthorblockA{\textit{School of Information and Physical Sciences} \\
        \textit{The University of Newcastle, University Drive}\\
        Callaghan, NSW 2308, Australia \\
        stephan.chalup@newcastle.edu.au}
    }
    

    \IEEEoverridecommandlockouts
    \IEEEpubid{\makebox[\columnwidth]
    {
    979-8-3503-7903-7/24/\$31.00 ©2024 IEEE 
    } 
    }
        
    \maketitle
    
    \begin{abstract}
        Understanding the topological characteristics of data is important to many areas of research. Recent work has demonstrated that synthetic 4D image-type data can be useful to train 4D convolutional neural network models to see topological features in these data. These models also appear to tolerate the use of image preprocessing techniques where existing topological data analysis techniques such as persistent homology do not. This paper investigates how methods from algebraic topology, combined with image processing techniques such as morphology, can be used to generate topologically sophisticated and diverse-looking 2-, 3-, and 4D image-type data with topological labels in simulation. These approaches are illustrated in 2D and 3D with the aim of providing a roadmap towards achieving this in 4D.
    \end{abstract}
    
    \begin{IEEEkeywords}
        Betti numbers, topology, manifold, persistent homology, image data generation.
    \end{IEEEkeywords}

    \section{Introduction}\label{sec:introduction}
        
        Understanding the topological characteristics of 2D and 3D data can be critical in application areas such as material science, where imaging methods such as magnetic resonance imaging (MRI) and computed tomography (CT) are often used to study materials~\cite{AlSahlaniEtAl2018, Dua20B}.
        Similarly, these characteristics can be a research consideration in the medical arena~\cite{Lou21}. 
        Methods for the topological analysis of 4D data become relevant when high-dimensional data would lose essential information if reduced to 3D or 2D, for example, when the residual variance in manifold learning~\cite{TenenbaumEtAl2000} indicates that the outcome of the dimensionality reduction process lies in~4D. 

        \IEEEpubidadjcol
        
        Homology is an algebraic topological concept used for studying objects such as topological spaces, and persistent homology is a computational approach to homology that can estimate topological properties such as Betti numbers, which are studied in homology. Betti numbers capture the essential structure given by the holes of a manifold or topological space~\cite{Ede10} and are a more fine-grained signature than the more broadly known Euler characteristic $\chi$. Informally, the $n$\textsuperscript{th} Betti number $\beta_n$ can be interpreted as the number of $n$-dimensional holes in a space, although, $\beta_0$ is often thought of as the number of connected components;
        this discrepancy is rectified with `reduced' homology, which says that a point should have $\beta_0=0$~\cite[Chapter 2]{Hatcher2002}.
        
        Synthetic image-type data can be used to train convolutional neural networks (CNN) that perform topological data analysis (TDA)~\cite{Pau19,Han23}.
        Deformation algorithms and computer graphics software can be applied to produce more visually complex data in 2D and 3D. 
        Unfortunately, these options do not transfer readily to the 4D context because of computational issues and the lack of 4D compatible software.
        The primary aim of this paper is to address this issue through the generalisation of some 2D and 3D techniques that can afford the generation of topologically and geometrically diverse 4D image-type data. 
        The labels of these data will correlate with Betti numbers.

        \IEEEpubidadjcol
        
    \section{Geometric diversity in 2D and 3D topological data generation}\label{sec:geometric_diversity_in_2d_and_3d}
        Solutions to introducing geometric variety in 2D and 3D data while preserving topological properties have emerged from the application of pixel-wise deformation approaches, linear transformations, and non-parametric transformations (many of these, including elastic transforms, and spline- or Bayesian-based transforms, are reviewed in~\cite{Gla98, Mon01}), as well as capitalising on hardware, algorithmic, and machine learning advances.
        Historically, these solutions have generally emerged and been applied in the 2D setting first, and have then been extended to the 3D setting as relevant theory and hardware have been developed to enable this.
    
        Digital topology, and notions of connectivity and adjacency in binary images~\cite{Ros70, Ros74}, have been used in developing topology-preserving deformations~\cite{Kon89, Ros98} that are based on moving `simple' pixels, which are pixels whose deletion does not affect an image's topology. The application of digital topology in the 3D setting is discussed in~\cite{Kon89}.

        Elastic transformations~\cite{Gab96} have been used in computer vision experiments to good effect, where it has been observed that applying them to the MNIST dataset can produce improved results versus when using affine distortions~\cite{Sim03}.
        Smooth deformation fields can be generated by placing a random displacement vector on each component of a grid and then interpolating them to generate a dense full-resolution deformation field~\cite[Supplementary Figure 3]{Fal19}.
        Folds in the resulting point-cloud must be prevented when applying these deformations. Some strategies to alleviate this issue include minimising the difference between the displacement of neighbouring points~\cite{Mot02} and composing the original warp with subsequent warps~\cite{Sch06,Tid01,Yu12}.
        
        An unsupervised learning approach to deform 2D samples is demonstrated in~\cite{Bal20}, which depends only on performing continuous and topology-preserving deformations, together with interpolation and checking that continuity is preserved at each step.
        This work also highlights what guardrails may be needed to combat the reliability limitations that probabilistic machine learning architectures would demonstrate when applying topology-preserving deformations; even state-of-the-art transformer architectures~\cite{Vas17,Rad18}, such as large language models and generative pre-trained transformers (GPTs), are not immune~\cite{Zho23,Lec23}.
        
        Basic 2D and 3D topological point-cloud data was used in~\cite{Pau19}. The 2D samples comprised of multiple homological components (islands), and 3D samples with cavities that were derived from 3-balls and multi-holed donuts.
        In a later project, it was shown by~\cite{Pee23} that objects in 3D samples that had been deformed with a repulsive algorithm~\cite{Yu21B} could still be segmented by a neural network; unfortunately, this algorithm is already computationally expensive in 3D and can produce anomalies such as self-intersections.
        Neural style transfer software uses neural networks to manipulate images to adopt the visual style of another image but can demonstrate artifacts that could pose a similar problem~\cite{Zhu20,Qia21,Xia21}.

    \section{Visualising some 3-manifolds}\label{sec:visualising_some_3-manifolds}
        New topological spaces can be constructed from existing spaces. 
        Quotient spaces involve defining an equivalence relation on a topological space, which identifies points that are mapped to the same equivalence class to one another. 
        Product spaces are a Cartesian product space of two topological spaces equipped with a topology. 
        An example of a quotient space is an interval $I$ with its boundary points identified to one another, which describes a circle $S^1$ (Figure~\ref{fig:quotent_space_of_interval}).
        An example of a product space is the torus $S^1 \times S^1$. 
        The torus is also an example of a quotient space (Figure~\ref{fig:square_to_torus}).

        The exercise of describing 1- and 2-manifolds with intervals and squares can be extended to the 3-manifolds that are considered in this paper by regarding them as quotient spaces of the cube $I^3$; the equivalence relations usually involve identifying pairs of faces of the cube (Figure~\ref{fig:cube_depiction_of-3-manifold}). 
        Similarly as in the cases of the 1- and 2-sphere, the 3-sphere is homeomorphic to $I^3$ with its boundary identified to a point.
        To construct $S^1 \times S^2$ or $S^1 \times S^1 \times S^1$, the FRONT and BACK faces of a cube are identified to produce a square donut $S^1 \times I^2$ (Figure~\ref{fig:visualising_3d_manifolds_S1S2}).
        An appropriate relation on the $I^2$ factor (for the sphere or torus) will produce the required spaces. Another description of $S^1 \times S^1 \times S^1$ is to identify the opposite faces of the cube to one another.
        For $S^2 \times S^1$, the BACK face is spherically wrapped around the FRONT face by identifying the remaining faces to one interval; this produces a space that is homeomorphic to $S^2 \times I$ (Figure~\ref{fig:visualising_3d_manifolds_S2S1}). From here, the inner and outer boundary spheres are identified with one another, that is, the boundary points of each interval are identified to form circles (Figure~\ref{fig:quotent_space_of_interval}).

        \begin{figure}[ht]
            \centering
            \begin{tikzpicture}[line cap=round,line join=round,>=triangle 45,x=1.0cm,y=1.0cm,scale=0.8]
                \clip(-4.5,-1.5) rectangle (4.5,1.5);
                \draw [line width=0.5pt] (-4.,0.)-- (-1.5,0.);
                \draw [line width=0.5pt] (3.,0.) circle (1.cm);
                
                \draw [line width=2pt, , -to] (-0.1,0.)-- (.6,0.);
                
                \begin{scriptsize}
                    \draw [fill=black] (-4.,0.) circle (1pt);
                    \draw [fill=black] (-1.5,0.) circle (1pt);
                    \draw [fill=black] (4.,0.) circle (1pt);
        
                    \draw[color=black] (-2.75,.2) node {$I$};
                    \draw[color=black] (3.,0.) node {$S^1$};
                \end{scriptsize}
            \end{tikzpicture}
            \caption[The quotient space of the interval]{The interval under the equivalence relation that identifies its boundary points is homeomorphic to a~circle.}
            \label{fig:quotent_space_of_interval}
        \end{figure}
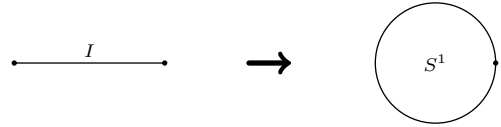

        \begin{figure}[ht]
            \centering
            
            \tdplotsetmaincoords{70}{0}
            \tikzset{declare function={
                torusx(\u,\v,\R,\r)=cos(\u)*(\R + \r*cos(\v)); 
                torusy(\u,\v,\R,\r)=(\R + \r*cos(\v))*sin(\u);
                torusz(\u,\v,\R,\r)=\r*sin(\v);
                vcrit1(\u,\th)=atan(tan(\th)*sin(\u));
                vcrit2(\u,\th)=180+atan(tan(\th)*sin(\u));
                disc(\th,\R,\r)=((pow(\r,2)-pow(\R,2))*pow(cot(\th),2)+%
                pow(\r,2)*(2+pow(tan(\th),2)))/pow(\R,2);
                umax(\th,\R,\r)=ifthenelse(disc(\th,\R,\r)>0,asin(sqrt(abs(disc(\th,\R,\r)))),0);
                }
            }
            
            \begin{tikzpicture}[tdplot_main_coords,scale=0.7]
                
                \fill[line width=2.pt,color=gray,fill=black,fill opacity=0.1] (-5.8,-3.4) -- (-5.8,3.4) -- (-8.2,3.4) -- (-8.2,-3.4) -- cycle;
                
                \draw [line width=0.5pt,-<] (-5.8,-3.4)-- (-5.8,0);
                \draw [line width=0.5pt] (-5.8,0)-- (-5.8,3.4);
                
                \draw [line width=0.5pt,-<] (-8.2,-3.4)-- (-8.2,0);
                \draw [line width=0.5pt] (-8.2,0)-- (-8.2,3.4);
                
                \draw [line width=0.5pt,-<<] (-5.8,3.4)-- (-7.,3.4);
                \draw [line width=0.5pt] (-7.,3.4)-- (-8.2,3.4);
                
                \draw [line width=0.5pt,-<<] (-5.8,-3.4)-- (-7.,-3.4);
                \draw [line width=0.5pt] (-7.,-3.4)-- (-8.2,-3.4);

                \draw [line width=2pt, ->, color=black] (-3.8,0)-- (-3.1,0);
            
                    
                \pgfmathsetmacro{\R}{1.2} 
                \pgfmathsetmacro{\r}{0.3} 

                \draw[thin,fill=black,even odd rule,fill opacity=0.1] plot[variable=\x,domain=0:360,smooth,samples=71]
                ({torusx(\x,vcrit1(\x,\tdplotmaintheta),\R,\r)},
                {torusy(\x,vcrit1(\x,\tdplotmaintheta),\R,\r)},
                {torusz(\x,vcrit1(\x,\tdplotmaintheta),\R,\r)}) 
                plot[variable=\x,
                domain={-180+umax(\tdplotmaintheta,\R,\r)}:{-umax(\tdplotmaintheta,\R,\r)},smooth,samples=51]
                ({torusx(\x,vcrit2(\x,\tdplotmaintheta),\R,\r)},
                {torusy(\x,vcrit2(\x,\tdplotmaintheta),\R,\r)},
                {torusz(\x,vcrit2(\x,\tdplotmaintheta),\R,\r)})
                plot[variable=\x,
                domain={umax(\tdplotmaintheta,\R,\r)}:{180-umax(\tdplotmaintheta,\R,\r)},smooth,samples=51]
                ({torusx(\x,vcrit2(\x,\tdplotmaintheta),\R,\r)},
                {torusy(\x,vcrit2(\x,\tdplotmaintheta),\R,\r)},
                {torusz(\x,vcrit2(\x,\tdplotmaintheta),\R,\r)});
                
                \draw[thin] plot[variable=\x,
                domain={-180+umax(\tdplotmaintheta,\R,\r)/2}:{-umax(\tdplotmaintheta,\R,\r)/2},smooth,samples=51]
                ({torusx(\x,vcrit2(\x,\tdplotmaintheta),\R,\r)},
                {torusy(\x,vcrit2(\x,\tdplotmaintheta),\R,\r)},
                {torusz(\x,vcrit2(\x,\tdplotmaintheta),\R,\r)});
                
                \foreach \X  in {240}
                {
                    \draw[thin,dashed] 
                    plot[smooth,variable=\x,domain={360+vcrit1(\X,\tdplotmaintheta)}:{vcrit2(\X,\tdplotmaintheta)},samples=71]   
                    ({torusx(\X,\x,\R,\r)},{torusy(\X,\x,\R,\r)},{torusz(\X,\x,\R,\r)});
                
                    \draw[thin] 
                    plot[smooth,variable=\x,domain={vcrit2(\X,\tdplotmaintheta)}:{vcrit1(\X,\tdplotmaintheta)},samples=71]   
                    ({torusx(\X,\x,\R,\r)},{torusy(\X,\x,\R,\r)},{torusz(\X,\x,\R,\r)})
                    ;
                    
                    \draw[thin,->] 
                    plot[smooth,variable=\x,domain={vcrit2(\X,\tdplotmaintheta)}:{vcrit1(\X,\tdplotmaintheta)+90},samples=71]   
                    ({torusx(\X,\x,\R,\r)},{torusy(\X,\x,\R,\r)},{torusz(\X,\x,\R,\r)})
                   ;
                }
                
                
                \draw[thin,->>] plot[smooth,variable=\x,domain=0:360,samples=71]   
                ({torusx(\x,70,\R,\r)},
                {torusy(\x,70,\R,\r)},
                {torusz(\x,70,\R,\r)})
                ;
                
            \end{tikzpicture}
        
            \caption[Constructing a torus]{The matching arrows of the square can be identified to construct a torus.}
            \label{fig:square_to_torus}
        \end{figure}
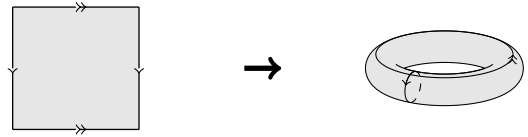

        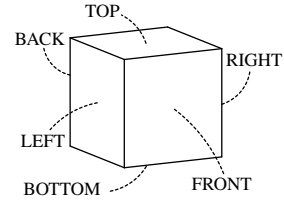
\begin{figure}[ht]
            \centering
            
            \begin{tikzpicture}[line cap=round,line join=round,>=triangle 45,x=0.8cm,y=0.8cm, scale=0.9]
                \clip(-2.2,-0.6) rectangle (3.,3.1);
                \draw [line width=.5pt] (0.,2.)-- (0.,0.);
                \draw [line width=.5pt] (0.,0.)-- (1.8,0.2);
                \draw [line width=.5pt] (1.8,0.2)-- (1.8,2.2);
                \draw [line width=.5pt] (1.8,2.2)-- (0.,2.);
                \draw [line width=.5pt] (0.,2.)-- (-1.,2.4);
                \draw [line width=.5pt] (0.,0.)-- (-1.,0.4);
                \draw [line width=.5pt] (-1.,0.4)-- (-1.,2.4);
                \draw [line width=.5pt] (1.8,2.2)-- (0.8,2.6);
                \draw [line width=.5pt] (0.8,2.6)-- (-1.,2.4);
                
                \draw [shift={(-0.20267381944419605,-0.329408692886877)},line width=.5pt,dash pattern=on 1pt off 1pt]  plot[domain=1.7628072710510483:2.4174735966045184,variable=\t]({1.*1.5580416578262108*cos(\t r)+0.*1.5580416578262108*sin(\t r)},{0.*1.5580416578262108*cos(\t r)+1.*1.5580416578262108*sin(\t r)});
                \draw [shift={(-0.8048881924248822,-1.182708412273108)},line width=.5pt,dash pattern=on 1pt off 1pt]  plot[domain=0.3863050381541621:0.9293009343585306,variable=\t]({1.*2.8491053761018024*cos(\t r)+0.*2.8491053761018024*sin(\t r)},{0.*2.8491053761018024*cos(\t r)+1.*2.8491053761018024*sin(\t r)});
                \draw [shift={(-0.466025320820407,2.0079144592029614)},line width=.5pt,dash pattern=on 1pt off 1pt]  plot[domain=0.3252905845369018:1.0999637210414173,variable=\t]({1.*0.9139550423542664*cos(\t r)+0.*0.9139550423542664*sin(\t r)},{0.*0.9139550423542664*cos(\t r)+1.*0.9139550423542664*sin(\t r)});
                \draw [shift={(-0.4438981410580716,0.547419985373011)},line width=.5pt,dash pattern=on 1pt off 1pt]  plot[domain=4.899488804782056:5.757496524619099,variable=\t]({1.*0.9980337960273997*cos(\t r)+0.*0.9980337960273997*sin(\t r)},{0.*0.9980337960273997*cos(\t r)+1.*0.9980337960273997*sin(\t r)});
                \draw [shift={(-0.8584294923102281,2.3374565510641974)},line width=.5pt,dash pattern=on 1pt off 1pt]  plot[domain=3.3494412187814135:4.489759578854905,variable=\t]({1.*0.6411856108698*cos(\t r)+0.*0.6411856108698*sin(\t r)},{0.*0.6411856108698*cos(\t r)+1.*0.6411856108698*sin(\t r)});
                \draw [shift={(1.5755220312892595,1.889787793688538)},line width=.5pt,dash pattern=on 1pt off 1pt]  plot[domain=5.015797523228928:6.132596497954475,variable=\t]({1.*0.7513283481993973*cos(\t r)+0.*0.7513283481993973*sin(\t r)},{0.*0.7513283481993973*cos(\t r)+1.*0.7513283481993973*sin(\t r)});
                
                \begin{scriptsize}
                    \draw[color=black] (-1.5,0.5) node {LEFT};
                    \draw[color=black] (1.8,-0.3) node {FRONT};
                    \draw[color=black] (-0.4,2.891928357472642) node {TOP};
                    \draw[color=black] (-1.15,-0.3834006342054088) node {BOTTOM};
                    \draw[color=black] (-1.559169697737729,2.3781512607388304) node {BACK};
                    \draw[color=black] (2.4,1.99) node {RIGHT};
                \end{scriptsize}
            \end{tikzpicture}
        
            \caption[The quotient space of a cube]{The 3-manifolds that are considered in this project can be defined by quotient spaces of the cube $I^3$. 
            }
            \label{fig:cube_depiction_of-3-manifold}
        \end{figure}

        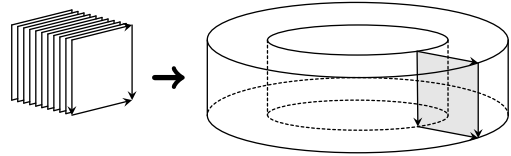
\begin{figure}[ht]
            \centering
            \begin{tikzpicture}[line cap=round,line join=round,>=triangle 45,x=1.0cm,y=1.0cm, scale=1.]
                \clip(-3.8,-0.7) rectangle (3.2,1.7);
        
                \fill[line width=2.pt,fill=black,fill opacity=0.1] (1.7886190751269053,0.8500904338842461) -- (2.6014646742502965,0.6955756181424648) -- (2.601545624953649,-0.30467076489246164) -- (1.799349589886422,-0.1520646350454237) -- cycle;
                
                
                \draw [line width=.5pt,dash pattern=on 1pt off 1pt] (3,0) arc(0:180:2.0001759842496933cm and 0.4999972506316682cm);
                \draw [line width=.5pt] (-1,0) arc(180:360:2.0001759842496933cm and 0.4999972506316682cm);
                
                \draw [rotate around={179.81401582324264:(1.,1.)},line width=.5pt] (1.,1.) ellipse (2.000158072783076cm and 0.49999753042384654cm);
                
                \draw [rotate around={-179.93729816154948:(1.,1.)},line width=.5pt] (1.,1.) ellipse (1.2000251506433999cm and 0.1999998835655981cm);
                \draw [rotate around={179.7884868375666:(1.,0.)},line width=.5pt,dash pattern=on 1pt off 1pt] (1.,0.) ellipse (1.200286295215612cm and 0.19999867504653676cm);
                
                \draw [line width=.5pt] (-1.,0.)-- (-1.,1.);
                \draw [line width=.5pt] (3.,0.)-- (3.,1.);
                
                \draw [line width=.5pt,dash pattern=on 1pt off 1pt] (-0.2,1.)-- (-0.2,0.);
                \draw [line width=.5pt,dash pattern=on 1pt off 1pt] (2.2,1.)-- (2.2,0.);
                \draw [line width=.5pt,-stealth] (2.6014646742502965,0.6955756181424648)-- (2.601545624953649,-0.30467076489246164);
                \draw [line width=.5pt,-stealth] (1.7886190751269053,0.8500904338842461)-- (1.799349589886422,-0.1520646350454237);
                \draw [line width=.5pt,-stealth] (1.7886190751269053,0.8500904338842461)-- (2.6014646742502965,0.6955756181424648);
                \draw [line width=.5pt,-stealth] (1.799349589886422,-0.1520646350454237)-- (2.601545624953649,-0.30467076489246164);

                \draw [line width=.5pt,stealth-] (-2.8,0.)-- (-2.8,1.);
                \draw [line width=.5pt,-stealth] (-2.8,1.)-- (-2.,1.2);
                \draw [line width=.5pt,-stealth] (-2.,1.2)-- (-2.,0.2);
                \draw [line width=.5pt,stealth-] (-2.,0.2)-- (-2.8,0.);
                
                \draw [line width=.5pt] (-2.96,1.04)-- (-2.96,0.04);
                \draw [line width=.5pt] (-2.88,1.02)-- (-2.88,0.02);
                \draw [line width=.5pt] (-2.88,1.02)-- (-2.08,1.22);
                \draw [line width=.5pt] (-2.96,1.04)-- (-2.16,1.24);
                \draw [line width=.5pt] (-2.88,0.02)-- (-2.8,0.04);
                \draw [line width=.5pt] (-2.96,0.04)-- (-2.88,0.06);
                \draw [line width=.5pt] (-2.08,1.22)-- (-2.08,1.18);
                \draw [line width=.5pt] (-2.16,1.2)-- (-2.16,1.24);
                \draw [line width=.5pt] (-3.04,0.06)-- (-3.04,1.06);
                \draw [line width=.5pt] (-3.12,1.08)-- (-3.12,0.08);
                \draw [line width=.5pt] (-3.2,0.1)-- (-3.2,1.1);
                \draw [line width=.5pt] (-3.28,1.12)-- (-3.28,0.12);
                \draw [line width=.5pt] (-3.36,0.14)-- (-3.36,1.14);
                \draw [line width=.5pt] (-3.44,1.16)-- (-3.44,0.16);
                \draw [line width=.5pt] (-3.52,0.18)-- (-3.52,1.18);
                \draw [line width=.5pt] (-3.6,1.2)-- (-3.6,0.2);
                \draw [line width=.5pt] (-3.04,0.06)-- (-2.96,0.08);
                \draw [line width=.5pt] (-3.04,0.1)-- (-3.12,0.08);
                \draw [line width=.5pt] (-3.12,0.12)-- (-3.2,0.1);
                \draw [line width=.5pt] (-3.2,0.14)-- (-3.28,0.12);
                \draw [line width=.5pt] (-3.28,0.16)-- (-3.36,0.14);
                \draw [line width=.5pt] (-3.36,0.18)-- (-3.44,0.16);
                \draw [line width=.5pt] (-3.44,0.2)-- (-3.52,0.18);
                \draw [line width=.5pt] (-3.52,0.22)-- (-3.6,0.2);
                \draw [line width=.5pt] (-3.04,1.06)-- (-2.24,1.26);
                \draw [line width=.5pt] (-2.32,1.28)-- (-3.12,1.08);
                \draw [line width=.5pt] (-3.2,1.1)-- (-2.4,1.3);
                \draw [line width=.5pt] (-2.48,1.32)-- (-3.28,1.12);
                \draw [line width=.5pt] (-3.36,1.14)-- (-2.56,1.34);
                \draw [line width=.5pt] (-2.64,1.36)-- (-3.44,1.16);
                \draw [line width=.5pt] (-3.52,1.18)-- (-2.72,1.38);
                \draw [line width=.5pt] (-2.8,1.4)-- (-3.6,1.2);
                \draw [line width=.5pt] (-2.24,1.22)-- (-2.24,1.26);
                \draw [line width=.5pt] (-2.32,1.24)-- (-2.32,1.28);
                \draw [line width=.5pt] (-2.4,1.26)-- (-2.4,1.3);
                \draw [line width=.5pt] (-2.48,1.28)-- (-2.48,1.32);
                \draw [line width=.5pt] (-2.56,1.3)-- (-2.56,1.34);
                \draw [line width=.5pt] (-2.64,1.32)-- (-2.64,1.36);
                \draw [line width=.5pt] (-2.72,1.34)-- (-2.72,1.38);
                \draw [line width=.5pt] (-2.8,1.36)-- (-2.8,1.4);
                \draw [line width=2.pt, , -to] (-1.7,0.5)-- (-1.3,0.5);
            \end{tikzpicture}
            \caption[Visualising $S^1 \times I^2$]{Visualising $S^1 \times I^2$. With an appropriate labelling of $I^2$,
            this donut can be used to understand $S^1 \times S^2$ and $S^1 \times S^1 \times S^1$.}
            \label{fig:visualising_3d_manifolds_S1S2}
        \end{figure}

        \begin{figure}[ht]
            \centering
            \begin{tikzpicture}[line cap=round,line join=round,>=triangle 45,x=1.0cm,y=1.0cm, scale=1.]
                \clip(-4.,-0.8) rectangle (4.5,2.);
                \draw [line width=.5pt] (-2.8,0.)-- (-3.6,0.2);
                \draw [line width=.5pt] (-2.8,1.)-- (-3.6,1.2);
                \draw [line width=.5pt] (-2.,1.2)-- (-2.8,1.4);
                \draw [line width=.5pt] (-3.6,1.1)-- (-2.8,0.9);
                \draw [line width=.5pt] (-2.8,0.8)-- (-3.6,1.);
                \draw [line width=.5pt] (-3.6,0.9)-- (-2.8,0.7);
                \draw [line width=.5pt] (-2.8,0.6)-- (-3.6,0.8);
                \draw [line width=.5pt] (-2.8,0.5)-- (-3.6,0.7);
                \draw [line width=.5pt] (-3.6,0.6)-- (-2.8,0.4);
                \draw [line width=.5pt] (-3.6,0.5)-- (-2.8,0.3);
                \draw [line width=.5pt] (-2.8,0.2)-- (-3.6,0.4);
                \draw [line width=.5pt] (-3.6,0.3)-- (-2.8,0.1);
                \draw [line width=.5pt] (-2.72,1.02)-- (-3.52,1.22);
                \draw [line width=.5pt] (-3.44,1.24)-- (-2.64,1.04);
                \draw [line width=.5pt] (-3.36,1.26)-- (-2.56,1.06);
                \draw [line width=.5pt] (-3.28,1.28)-- (-2.48,1.08);
                \draw [line width=.5pt] (-2.4,1.1)-- (-3.2,1.3);
                \draw [line width=.5pt] (-3.12,1.32)-- (-2.32,1.12);
                \draw [line width=.5pt] (-2.24,1.14)-- (-3.04,1.34);
                \draw [line width=.5pt] (-2.96,1.36)-- (-2.16,1.16);
                \draw [line width=.5pt] (-2.08,1.18)-- (-2.88,1.38);
                \draw [rotate around={89.91280081574142:(0.,0.5)},line width=.5pt] (0.,0.5) ellipse (0.5000030400766172cm and 0.1999998054371549cm);
                \draw [rotate around={-89.96531313300922:(0.,0.5)},line width=.5pt] (0.,0.5) ellipse (1.1000007740687285cm and 0.499999927303931cm);
                \draw [shift={(0.105,0.5)},line width=.5pt]  plot[domain=1.6659625333488637:4.617222773830722,variable=\t]({1.*1.105*cos(\t r)+0.*1.105*sin(\t r)},{0.*1.105*cos(\t r)+1.*1.105*sin(\t r)});
                \draw [shift={(0.1125,0.5)},line width=.5pt,dash pattern=on 1pt off 1pt]  plot[domain=1.7921107691426879:4.491074538036898,variable=\t]({1.*0.5125*cos(\t r)+0.*0.5125*sin(\t r)},{0.*0.5125*cos(\t r)+1.*0.5125*sin(\t r)});
                \draw [line width=.5pt] (-0.12386135732897258,0.8915861352053773)-- (-0.28573545209642753,1.403349065362042);
                \draw [line width=.5pt] (0.12420246294302159,0.8928940459829107)-- (0.285340122748188,1.4026265071669806);
                \draw [line width=.5pt] (0.12511559369963354,0.11091757437266804)-- (0.28891106164701563,-0.39845089114993126);
                \draw [line width=.5pt] (-0.12448980667350135,0.10767388615515608)-- (-0.28606258676215823,-0.40151817228815767);
                \draw [shift={(-0.7120505358471763,1.1940345257679017)},line width=.5pt,dash pattern=on 1pt off 1pt]  plot[domain=3.030278258359347:4.69056704806345,variable=\t]({1.*0.2865757646825632*cos(\t r)+0.*0.2865757646825632*sin(\t r)},{0.*0.2865757646825632*cos(\t r)+1.*0.2865757646825632*sin(\t r)});
                \draw [shift={(0.048032461376761815,-0.239838029496015)},line width=.5pt,dash pattern=on 1pt off 1pt]  plot[domain=0.22293390811987085:1.6356282821693038,variable=\t]({1.*0.7413955942912369*cos(\t r)+0.*0.7413955942912369*sin(\t r)},{0.*0.7413955942912369*cos(\t r)+1.*0.7413955942912369*sin(\t r)});
                \draw [shift={(0.4773247002220777,1.2708412798175703)},line width=.5pt,dash pattern=on 1pt off 1pt]  plot[domain=1.647693955946872:3.1283677712920865,variable=\t]({1.*0.4680311228219401*cos(\t r)+0.*0.4680311228219401*sin(\t r)},{0.*0.4680311228219401*cos(\t r)+1.*0.4680311228219401*sin(\t r)});
                \draw [shift={(0.2965751211334874,1.201083301170725)},line width=.5pt,dash pattern=on 1pt off 1pt]  plot[domain=4.8878212180645555:5.9940545748340845,variable=\t]({1.*0.47133087616014446*cos(\t r)+0.*0.47133087616014446*sin(\t r)},{0.*0.47133087616014446*cos(\t r)+1.*0.47133087616014446*sin(\t r)});
                \draw [line width=.5pt] (3.,0.5) circle (1.1cm);
                \draw [line width=.5pt,dash pattern=on 1pt off 1pt] (3.,0.5) circle (0.5cm);
                \draw [rotate around={179.8141055644964:(3.,0.5)},line width=.5pt,dash pattern=on 1pt off 1pt] (3.,0.5) ellipse (0.5000631725903781cm and 0.09999949471886972cm);
                \draw [rotate around={179.53048733823593:(3.,0.5)},line width=.5pt,dash pattern=on 1pt off 1pt] (3.,0.5) ellipse (1.100678789592313cm and 0.24999203923959545cm);
                \draw [line width=.5pt] (3.5,0.5)-- (4.1,0.5);
                
                \draw [line width=2.pt, , -to] (-1.7,0.5)-- (-1.3,0.5);
                \draw [line width=2.pt, , -to] (1.1,0.5)-- (1.5,0.5);
                \begin{scriptsize}
            
                    \draw [fill=black] (-2.,0.2) circle (0.25pt);
                    \draw [fill=black] (-2.,1.1) circle (0.25pt);
                    \draw [fill=black] (-2.,1.) circle (0.25pt);
                    \draw [fill=black] (-2.,0.9) circle (0.25pt);
                    \draw [fill=black] (-2.,0.8) circle (0.25pt);
                    \draw [fill=black] (-2.,0.7) circle (0.25pt);
                    \draw [fill=black] (-2.,0.6) circle (0.25pt);
                    \draw [fill=black] (-2.,0.5) circle (0.25pt);
                    \draw [fill=black] (-2.,0.4) circle (0.25pt);
                    \draw [fill=black] (-2.,0.3) circle (0.25pt);
            
                    \draw [fill=black] (-2.72,0.02) circle (0.25pt);
                    \draw [fill=black] (-2.64,0.04) circle (0.25pt);
                    \draw [fill=black] (-2.56,0.06) circle (0.25pt);
                    \draw [fill=black] (-2.48,0.08) circle (0.25pt);
                    \draw [fill=black] (-2.4,0.1) circle (0.25pt);
                    \draw [fill=black] (-2.32,0.12) circle (0.25pt);
                    \draw [fill=black] (-2.24,0.14) circle (0.25pt);
                    \draw [fill=black] (-2.16,0.16) circle (0.25pt);
                    \draw [fill=black] (-2.08,0.18) circle (0.25pt);
            
                    \draw [fill=black] (-2.72,0.92) circle (0.25pt);
                    \draw [fill=black] (-2.64,0.94) circle (0.25pt);
                    \draw [fill=black] (-2.56,0.96) circle (0.25pt);
                    \draw [fill=black] (-2.48,0.98) circle (0.25pt);
                    \draw [fill=black] (-2.4,1.) circle (0.25pt);
                    \draw [fill=black] (-2.32,1.02) circle (0.25pt);
                    \draw [fill=black] (-2.24,1.04) circle (0.25pt);
                    \draw [fill=black] (-2.16,1.06) circle (0.25pt);
                    \draw [fill=black] (-2.08,1.08) circle (0.25pt);
                    \draw [fill=black] (-2.08,0.98) circle (0.25pt);
                    \draw [fill=black] (-2.16,0.96) circle (0.25pt);
                    \draw [fill=black] (-2.24,0.94) circle (0.25pt);
                    \draw [fill=black] (-2.32,0.92) circle (0.25pt);
                    \draw [fill=black] (-2.4,0.9) circle (0.25pt);
                    \draw [fill=black] (-2.48,0.88) circle (0.25pt);
                    \draw [fill=black] (-2.56,0.86) circle (0.25pt);
                    \draw [fill=black] (-2.64,0.84) circle (0.25pt);
                    \draw [fill=black] (-2.72,0.82) circle (0.25pt);
                    \draw [fill=black] (-2.72,0.72) circle (0.25pt);
                    \draw [fill=black] (-2.72,0.62) circle (0.25pt);
                    \draw [fill=black] (-2.72,0.52) circle (0.25pt);
                    \draw [fill=black] (-2.72,0.42) circle (0.25pt);
                    \draw [fill=black] (-2.72,0.32) circle (0.25pt);
                    \draw [fill=black] (-2.72,0.22) circle (0.25pt);
                    \draw [fill=black] (-2.72,0.12) circle (0.25pt);
                    \draw [fill=black] (-2.64,0.14) circle (0.25pt);
                    \draw [fill=black] (-2.64,0.24) circle (0.25pt);
                    \draw [fill=black] (-2.64,0.34) circle (0.25pt);
                    \draw [fill=black] (-2.64,0.44) circle (0.25pt);
                    \draw [fill=black] (-2.64,0.54) circle (0.25pt);
                    \draw [fill=black] (-2.64,0.64) circle (0.25pt);
                    \draw [fill=black] (-2.64,0.74) circle (0.25pt);
                    \draw [fill=black] (-2.56,0.76) circle (0.25pt);
                    \draw [fill=black] (-2.56,0.66) circle (0.25pt);
                    \draw [fill=black] (-2.48,0.78) circle (0.25pt);
                    \draw [fill=black] (-2.4,0.8) circle (0.25pt);
                    \draw [fill=black] (-2.32,0.82) circle (0.25pt);
                    \draw [fill=black] (-2.24,0.84) circle (0.25pt);
                    \draw [fill=black] (-2.16,0.86) circle (0.25pt);
                    \draw [fill=black] (-2.08,0.88) circle (0.25pt);
                    \draw [fill=black] (-2.08,0.78) circle (0.25pt);
                    \draw [fill=black] (-2.16,0.76) circle (0.25pt);
                    \draw [fill=black] (-2.24,0.74) circle (0.25pt);
                    \draw [fill=black] (-2.32,0.72) circle (0.25pt);
                    \draw [fill=black] (-2.4,0.7) circle (0.25pt);
                    \draw [fill=black] (-2.48,0.68) circle (0.25pt);
                    \draw [fill=black] (-2.56,0.56) circle (0.25pt);
                    \draw [fill=black] (-2.48,0.58) circle (0.25pt);
                    \draw [fill=black] (-2.4,0.6) circle (0.25pt);
                    \draw [fill=black] (-2.32,0.62) circle (0.25pt);
                    \draw [fill=black] (-2.24,0.64) circle (0.25pt);
                    \draw [fill=black] (-2.16,0.66) circle (0.25pt);
                    \draw [fill=black] (-2.08,0.68) circle (0.25pt);
                    \draw [fill=black] (-2.56,0.46) circle (0.25pt);
                    \draw [fill=black] (-2.56,0.36) circle (0.25pt);
                    \draw [fill=black] (-2.56,0.26) circle (0.25pt);
                    \draw [fill=black] (-2.56,0.16) circle (0.25pt);
                    \draw [fill=black] (-2.48,0.18) circle (0.25pt);
                    \draw [fill=black] (-2.48,0.28) circle (0.25pt);
                    \draw [fill=black] (-2.48,0.38) circle (0.25pt);
                    \draw [fill=black] (-2.48,0.48) circle (0.25pt);
                    \draw [fill=black] (-2.4,0.5) circle (0.25pt);
                    \draw [fill=black] (-2.32,0.52) circle (0.25pt);
                    \draw [fill=black] (-2.24,0.54) circle (0.25pt);
                    \draw [fill=black] (-2.16,0.56) circle (0.25pt);
                    \draw [fill=black] (-2.08,0.58) circle (0.25pt);
                    \draw [fill=black] (-2.08,0.48) circle (0.25pt);
                    \draw [fill=black] (-2.16,0.46) circle (0.25pt);
                    \draw [fill=black] (-2.24,0.44) circle (0.25pt);
                    \draw [fill=black] (-2.32,0.42) circle (0.25pt);
                    \draw [fill=black] (-2.4,0.4) circle (0.25pt);
                    \draw [fill=black] (-2.4,0.3) circle (0.25pt);
                    \draw [fill=black] (-2.4,0.2) circle (0.25pt);
                    \draw [fill=black] (-2.32,0.22) circle (0.25pt);
                    \draw [fill=black] (-2.32,0.32) circle (0.25pt);
                    \draw [fill=black] (-2.24,0.34) circle (0.25pt);
                    \draw [fill=black] (-2.16,0.36) circle (0.25pt);
                    \draw [fill=black] (-2.08,0.38) circle (0.25pt);
                    \draw [fill=black] (-2.08,0.28) circle (0.25pt);
                    \draw [fill=black] (-2.16,0.26) circle (0.25pt);
                    \draw [fill=black] (-2.24,0.24) circle (0.25pt);
            
                    \draw[color=black] (-1.1,1.4) node {BACK};
                    \draw[color=black] (1.0,-0.2) node {FRONT};
                    \draw[color=black] (0.8,1.7) node {TOP};
                    \draw[color=black] (1.,1.2) node {RIGHT};
            
                    \draw [fill=black] (3.5,0.5) circle (1.pt);
                    \draw [fill=black] (4.1,0.5) circle (1.0pt);
            
                \end{scriptsize}
            \end{tikzpicture}
        
            \caption[Visualising $S^2 \times S^1$]{Visualising $S^2 \times S^1$. The BACK face is spherically wrapped around the FRONT face to give $S^2 \times I$. The inner and outer boundary spheres are then identified with one another.}
            \label{fig:visualising_3d_manifolds_S2S1}
        \end{figure}
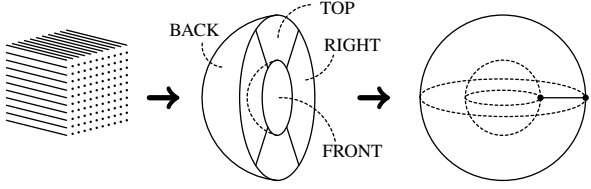

    \section{Topological data diversity in 4D}\label{sec:topological_diversity_in_4D}
        The data representation that is used in this paper follows the work of~\cite{Pau19,Han23}, in which
        each sample began as a tensor with each dimension having an equal length $I$ and every entry set to~1; this would represent a solid square $I^2$, cube $I^3$, or higher-dimensional cube analogue. Cavities were then introduced into the interior of a sample by setting the pixels that represented a cavity to 0. 
        In~\cite{Han23,Han23arXiv}, all cavities had a canonical appearance and were generated by using implicit equations.
        Increasing the topological and geometric diversity of synthetic 4D image-type data could be achieved by considering a wider range of manifolds, generating multi-component samples (where $\beta_0 > 1$), and applying topology-preserving deformations.
        
        \subsection{Links and knots}\label{subsec:links_and_knots}
            Mathematical links and knots are the subject of an area of study known as knot theory~\cite{Van71}. A knot is an embedding of $S^1$ in $\mathbb{R}^3$. A non-trivial example of a knot is the trefoil knot~\cite[Chapter 1.1]{Ada04}. A non-trivial example of a link is the Hopf link, which consists of two circles that are knotted together like a chain link~\cite{Kus98}. Both concepts could be used to generate data that contain linked and knotted features by
            linking and knotting an $S^1$ factor of the manifolds that are described in Section~\ref{sec:visualising_some_3-manifolds}.
            For example, 
            crossing a Hopf link with a circle
            could be used to generate two tori that are linked together
            and
            a knotted donut could be generated by crossing a knot with $B^2$.
            In the 4D setting, these examples would no longer be linked or knotted, however, it would be possible to observe similar phenomena with knotted and linked 2-spheres~\cite{Art25,Van28,And59}, as well as with 3-manifolds and 4-manifolds-with-boundary, for example, by crossing a link or knot with $S^2$ or $B^3$. Linked and knotted spheres are illustrated in~\cite{Boe03}.

            Computing the homology of links and knots is straightforward because the constituent manifolds are independent components. The homology of a cube with these objects cut out from its interior can be computed by applying the same approach that was used in
            ~\cite{Han23,Han23arXiv}.

        \subsection{Sums}\label{subsec:sums}
            
            The connected sum is an operation that joins two manifolds together, and 
                is signified with an octothorpe (hash) symbol and an optional subscript $\#_{g}$ to denote $g$ repeated applications.
                For two disjoint and connected $n$-manifolds $M_1$ and $M_2$, their connected sum $M_1\#M_2$ is a connected $n$-manifold that is formed by deleting the interior of an $n$-ball from each manifold and identifying the resulting boundary $(n-1)$-spheres together. 
                For example, the connected sum of $g$ tori is a $g$-holed torus.
                
                The boundary connected sum extends the idea of a connected sum to manifolds with a non-empty boundary, where two $n$-manifolds-with-boundary are glued along an $(n-1)$-ball in their respective boundaries.
                The boundary connected sum is signified by  
                a musical natural symbol $\natural_{g}$, with an optional subscript to denote $g$ repeated applications.
                
                    The wedge sum of two topological spaces $X$ and $Y$ 
                    is a new space that arises by gluing $X$ and $Y$ at one point; a formal definition is provided by~\cite[Chapter 0]{Hat02}. For example, the wedge sum of two circles is homeomorphic to a `figure-eight' space. 
                    
                    The boundary connected sum of two oriented, closed, and connected $n$-dimensional manifolds-with-boundary has the homotopy type of their wedge sum~\cite[page 98 exercise]{Kos93}, and
                    the reduced homology groups of a wedge sum of spaces is isomorphic to the direct sum of the reduced homology groups of the constituent spaces~\cite[Corollary 2.25]{Hat02}. These properties afford the derivation of the Betti number labels of our data.
            
            \subsubsection{Computing homology groups and Betti numbers of connected sums}\label{subsubsec:computing_homology_groups_and_betti_numbers_of_connected_sums}

                Derivations of the results in this section follow~\cite{Han23}.
                Four 3-manifolds-with-boundary were considered in
                ~\cite{Han23,Han23arXiv}, 
                namely, $B^4$, $S^1 \times B^3$, $S^2 \times B^2$, and $S^1 \times S^1 \times B^2$.
                In the context of the boundary connected sum, the 4-ball $B^4$ corresponds to the identity element. 
                Combining two of the remaining three options (with repetitions allowed) means that there are six
                pairs of manifolds that result in a non-trivial boundary connected sum, 
                namely, 
                $( S^1 \times B^3 ) \natural ( S^1 \times B^3 )$, 
                $( S^2 \times B^2 ) \natural ( S^2 \times B^2 )$,
                $( S^1 \times S^1 \times B^2 ) \natural ( S^1 \times S^1 \times B^2 )$, 
                $( S^1 \times B^3 ) \natural ( S^1 \times S^1 \times B^2 )$, 
                $( S^2 \times B^2 ) \natural ( S^1 \times S^1 \times B^2 )$, and 
                $( S^1 \times B^3 ) \natural ( S^2 \times B^2 )$.
                The homology of the space that is produced by repeated applications of the (boundary) connected sum can be found by using an induction proof. 
                Table~\ref{tab:betti_numbers_connected_sum} summarises the results. 
                
                A 4D analogue of a tunnel can be produced by replacing an $S^1$ (or $S^2$) factor of a 4-manifold-with-boundary with an interval $I$ (or a plane $I^2$, respectively) and then cutting the resulting object out of the interior of a 4D cube.
                Note that $I \times B^3$ and $I^2 \times B^2$ are homeomorphic to $B^4$. Similarly, $I \times S^1 \times B^2$ is homeomorphic to $S^1 \times B^3$ and, more generally, $I \times \natural_g (S^1 \times B^2)$ is homeomorphic to $\natural_g (S^1 \times B^3)$.
                Hence, the descriptors of these objects are not explicitly listed.

                \begin{table*}[ht!]
                    \caption[Topology of connected sums]{Topology of connected sums.
                    The subscripts $g$, $h$, $i$ and $j$ that are seen in this table denote the number of copies of an object that are glued.
                    One should assume that $g+h+i>0$, and that $j>0$ when $i>0$.
                    Also note that while $S^1 \times B^3$ and $S^2 \times B^2$ have homeomorphic boundaries, the convention that is used in this paper is that $S^1 \times S^2$ is the boundary of $S^1 \times B^3$ and $S^2 \times S^1$ is the boundary of $S^2 \times B^2$.
                    }
                    \label{tab:betti_numbers_connected_sum}
                    \centering
                    \begin{adjustbox}{max width=\textwidth}
                    \renewcommand{\arraystretch}{1.3}
                    \begin{tabular}{lccccc}
                        \hline 
                        \multicolumn{1}{c}{Manifold} & $\beta_0$ & $\beta_1$ & $\beta_2$ & $\beta_3$ & $\chi$\\
                        \hline

                        $\#_g (S^1 \times S^2) \#_h (S^2 \times S^1) \#_i (S^1 \times \#_j (S^1 \times S^1))$ & 1 & $g+h+i(2j+1)$ & $g+h+i(2j+1)$ & 1 & 0\\
                        $\natural_g (S^1 \times B^3) \natural_h (S^2 \times B^2) \natural_i (S^1 \times \natural_j (S^1 \times B^2))$ & 1 & $g + i(j+1)$ & $h +ij$ & 0 & $1-g+h-i$\\
                        $I^4 - \natural_g (S^1 \times B^3) \natural_h (S^2 \times B^2) \natural_i (S^1 \times \natural_j (S^1 \times B^2))$ & 1 & $h+ij$ & $g+i(j+1)$ & 1 & $g-h+i$\\
                        \hline
                        
                    \end{tabular}
                    \end{adjustbox}
                \end{table*}

            \subsubsection{Visualising connected sums}\label{subsubsec:visualising_connected_sums}

                    The connected sum of two circles can be described by using two intervals, removing the interior of a 1-ball from each, and then identifying the boundaries of the resulting holes. 
                
                    2-manifolds such as the sphere, torus, projective plane, and Klein bottle can be constructed as a quotient space of the square $I^2$.
                    A connected sum of $n$ of these objects can be constructed with a $4n$-gon (equipped with an appropriate relation), which can be found by identifying each square at a corner (Figure~\ref{fig:connected_sum_of_2-manifolds}).
                    Since the connected sum is operating on closed and connected spaces, the location of the interior of the disc that is removed from $I^2$ to allow a gluing is a matter of perspective; Figure~\ref{fig:connected_sum_of_2-manifolds} illustrates this by shifting the identified boundaries to the bottom-right corner. For example, 
                    after constructing a connected sum of two spheres from the starting configuration in Figure~\ref{fig:connected_sum_of_2-manifolds}, and then cutting the resulting manifold open along the solid line that is shown in Figure~\ref{fig:visualising_a_connected_sum_of_spheres}, a space that is homeomorphic to the octagon that is shown right-most in Figure~\ref{fig:connected_sum_of_2-manifolds} is produced.

                    The 3-manifolds that have been used in this project can be described with quotient spaces of cubes (Section~\ref{sec:visualising_some_3-manifolds}).
                    Recall that $S^1 \times S^1 \times S^1$ can be described by identifying the opposite faces of the cube to one another. 
                    More generally, the product of an interval $I$ with a $4g$-gon (this results in a $(4g+2)$-hedron) along with an equivalence relation can be used to produce $S^1 \times \#_g (S^1 \times S^1)$. For example, an octagonal prism decahedron (alternatively, imagine $I \times \mathrm{octagon}$) can be used to represent a two-holed torus that has been rotated around a circle $S^1 \times \#_2 (S^1 \times S^1)$.
                    The connected sum of two 3-manifolds is shown in Figure~\ref{fig:connected_sum_of_3-manifolds}, where the triangular faces represent the boundaries of the balls that have been removed near the corner of each cube (or polyhedron) for gluing.
                    A connected sum of $n$ objects from the set $\{S^3, S^1 \times S^2, S^2 \times S^1\}$ and $m$ objects from the set $\{S^1 \times \#_g (S^1 \times S^1)\}_{g \in \mathbb{N}}$ can then be thought of as a quotient space of a 
                    $(6n + \sum_{i=1}^{m} (4g_i+2) )$-hedron
                    under an appropriate equivalence relation.

                    Finally, the boundary of the tunnels that were described at the end of Section~\ref{subsubsec:computing_homology_groups_and_betti_numbers_of_connected_sums} can be considered with a similar strategy to that which was described in Section~\ref{sec:visualising_some_3-manifolds} by leaving a pair of opposing faces unidentified.
                    \begin{figure}[ht]
                        \centering
                        \begin{tikzpicture}[line cap=round,line join=round,>=triangle 45,x=.4cm,y=.4cm,scale=0.65]
                            \clip(-14.,-4.5) rectangle (20.,5.5);
                            \draw [line width=.5pt] (-13.,2.)-- (-10.,3.);
                            \draw [line width=.5pt] (-13.,-2.)-- (-10.,-1.);
                            \draw [shift={(-16.75,0.)},line width=.5pt]  plot[domain=-0.48995732625372845:0.4899573262537283,variable=\t]({1.*4.25*cos(\t r)+0.*4.25*sin(\t r)},{0.*4.25*cos(\t r)+1.*4.25*sin(\t r)});
                            \draw [shift={(-13.75,1.)},line width=.5pt]  plot[domain=-0.48995732625372845:0.4899573262537283,variable=\t]({1.*4.25*cos(\t r)+0.*4.25*sin(\t r)},{0.*4.25*cos(\t r)+1.*4.25*sin(\t r)});
                            \draw [line width=.5pt] (-12.,-2.)-- (-9.,-1.);
                            \draw [shift={(-5.25,1.)},line width=.5pt]  plot[domain=2.651635327336065:3.631549979843521,variable=\t]({1.*4.25*cos(\t r)+0.*4.25*sin(\t r)},{0.*4.25*cos(\t r)+1.*4.25*sin(\t r)});

                            \draw [line width=.5pt] (-9.86057289267821,2.7131423691072634)-- (-9.,3.);
                            \draw [shift={(-8.25000000000032,-1.671828067673361E-13)},line width=.5pt]  plot[domain=3.5564859939801385:3.631549979843522,variable=\t]({1.*4.249999999999639*cos(\t r)+0.*4.249999999999639*sin(\t r)},{0.*4.249999999999639*cos(\t r)+1.*4.249999999999639*sin(\t r)});
                            
                            \draw [rotate around={-155.7473315277534:(-10.995793221392617,0.4983775824195023)},line width=.5pt] (-10.995793221392617,0.4983775824195023) ellipse (0.15382471817986665cm and 0.08802320283316944cm);                        	
                            \draw [line width=.5pt] (-5.,-2.)-- (-2.,-1.);
                            \draw [line width=.5pt] (-6.,2.)-- (-3.,3.);
                            \draw [shift={(-11.537998611294821,-1.5094996528237048)},line width=.5pt]  plot[domain=-0.07488273078650742:0.5648400570402354,variable=\t]({1.*6.556372200529274*cos(\t r)+0.*6.556372200529274*sin(\t r)},{0.*6.556372200529274*cos(\t r)+1.*6.556372200529274*sin(\t r)});
                            \draw [shift={(-8.46384190079103,-0.49096047519775754)},line width=.5pt]  plot[domain=-0.07858965326173095:0.5685469795154591,variable=\t]({1.*6.483854822266818*cos(\t r)+0.*6.483854822266818*sin(\t r)},{0.*6.483854822266818*cos(\t r)+1.*6.483854822266818*sin(\t r)});
                            \draw [shift={(2.041174082693662,-1.145587041346831)},line width=.5pt]  plot[domain=2.250307788803843:3.1055823003741314,variable=\t]({1.*4.043795686387121*cos(\t r)+0.*4.043795686387121*sin(\t r)},{0.*4.043795686387121*cos(\t r)+1.*4.043795686387121*sin(\t r)});

                            \draw [line width=.5pt] (-2.4775668263198254,2.)-- (-0.5,2.);
                            \draw [line width=.5pt] (-2.,-1.)-- (-2.1009668017712233,-0.5936077509024581);
                            \draw [line width=.5pt] (-2.,-1.)-- (-2.434846576258427,-0.9535718827714732);
                            \draw [shift={(-2.3961234441732073,-0.6546641308929578)},line width=.5pt]  plot[domain=0.20398382781633953:4.583557731253791,variable=\t]({1.*0.30140558238287074*cos(\t r)+0.*0.30140558238287074*sin(\t r)},{0.*0.30140558238287074*cos(\t r)+1.*0.30140558238287074*sin(\t r)});
                            \draw [line width=.5pt] (2.5,2.)-- (5.5,3.);
                            \draw [line width=.5pt] (3.5,-2.)-- (6.2,-1.1);
                            \draw [shift={(-3.,-1.5)},line width=.5pt]  plot[domain=-0.0767718912697779:0.5667292175235064,variable=\t]({1.*6.519202405202649*cos(\t r)+0.*6.519202405202649*sin(\t r)},{0.*6.519202405202649*cos(\t r)+1.*6.519202405202649*sin(\t r)});
                            \draw [shift={(0.75,-0.25)},line width=.5pt]  plot[domain=-0.043450895391530686:0.6000502134017536,variable=\t]({1.*5.755432216610669*cos(\t r)+0.*5.755432216610669*sin(\t r)},{0.*5.755432216610669*cos(\t r)+1.*5.755432216610669*sin(\t r)});
                            \draw [line width=.5pt] (6.2,-1.1)-- (5.864105287717209,-0.9812635400117803);
                            \draw [line width=.5pt] (6.5,-0.5)-- (6.308321268565669,-0.2673366638474936);
                            \draw [shift={(6.018748924949298,-0.5823227722813025)},line width=.5pt]  plot[domain=0.8274103311510846:4.342586814402425,variable=\t]({1.*0.4278649210834573*cos(\t r)+0.*0.4278649210834573*sin(\t r)},{0.*0.4278649210834573*cos(\t r)+1.*0.4278649210834573*sin(\t r)});
                            \draw [line width=.5pt] (6.047405025104273,2.)-- (8.,2.);
                            \draw [shift={(9.75,-0.75)},line width=.5pt]  plot[domain=2.137525544318403:3.064820762320015,variable=\t]({1.*3.2596012026013246*cos(\t r)+0.*3.2596012026013246*sin(\t r)},{0.*3.2596012026013246*cos(\t r)+1.*3.2596012026013246*sin(\t r)});
                            
                            \draw [line width=.5pt] (11.011491723397759,4.4920100298570045)-- (11.011491723397759,-0.007989970143001696);
                            \draw [line width=.5pt] (11.011491723397759,4.4920100298570045)-- (15.511491723397757,4.4920100298570045);
                            \draw [line width=.5pt] (15.511491723397757,4.4920100298570045)-- (15.511491723397757,0.9920100298569969);
                            \draw [line width=.5pt] (11.011491723397759,-0.007989970143001696)-- (14.511491723397759,-0.007989970143001696);
                            \draw [line width=.5pt] (14.511491723397759,-3.507989970143003)-- (14.511491723397759,-0.007989970143001696);
                            \draw [line width=.5pt] (14.511491723397759,-3.507989970143003)-- (19.011491723397718,-3.507989970143003);
                            \draw [line width=.5pt] (19.011491723397718,-3.507989970143003)-- (19.011491723397718,0.9920100298569969);
                            \draw [line width=.5pt] (19.011491723397718,0.9920100298569969)-- (15.511491723397757,0.9920100298569969);
                            \draw [line width=.5pt,dash pattern=on 1pt off 2pt] (14.511491723397759,-0.007989970143001696)-- (15.511491723397757,0.9920100298569969);
                            \draw [line width=.5pt] (-8.,0.)-- (-7.,0.);
                            \draw [line width=.5pt] (0.5,0.)-- (1.5,0.);
                            \draw [line width=.5pt] (8.6,0.)-- (9.6,0.);

                            \draw [line width=2pt, , -to] (-8.,0.)-- (-7.,0.);
                            \draw [line width=2pt, , -to] (.5,0.)-- (1.5,0.);
                            \draw [line width=2pt, , -to] (8.6,0.)-- (9.6,0.);
                        \end{tikzpicture}
                        \caption[Connected sum of 2-manifolds]{The connected sum of 2-manifolds represented as the gluing of two planes; each plane is assumed to be equipped with an appropriate relation on its edges.
                        The interior of a 2-ball is removed from each plane and the respective circular boundaries are identified. 
                        The location of the gluing is a matter of perspective and this is shown in the figure by shifting the glued circles to the bottom corner of the planes.
                        A circle can be described as an interval with its endpoints identified. One can see this identification occurring as the circle finally meets the corner point (second image), which is also the point to which the boundary points of the cut circular boundary, an interval, are mapped (third image). This interval is represented with a dotted line in the last image.}
                        \label{fig:connected_sum_of_2-manifolds}
                    \end{figure}
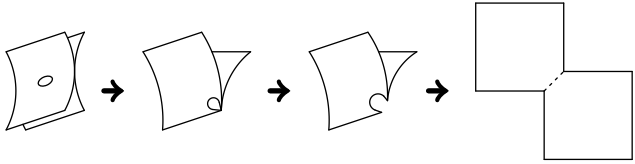

                    
                    \begin{figure}[ht]
                        \centering
                        \begin{tikzpicture}[line cap=round,line join=round,>=triangle 45,x=.3cm,y=.3cm, scale = 0.7]
                            \clip(-5.,-4.) rectangle (5.,4.);
                            \draw [shift={(2.68699106496609,-4.257358568999289)},line width=.5pt]  plot[domain=1.566605990687099:2.5937950104108536,variable=\t]({1.*4.062878456769053*cos(\t r)+0.*4.062878456769053*sin(\t r)},{0.*4.062878456769053*cos(\t r)+1.*4.062878456769053*sin(\t r)});
                            \draw [shift={(-2.3978975873716033,5.22324854865193)},line width=.5pt]  plot[domain=4.694536845337594:5.672764465543933,variable=\t]({1.*4.389204909298669*cos(\t r)+0.*4.389204909298669*sin(\t r)},{0.*4.389204909298669*cos(\t r)+1.*4.389204909298669*sin(\t r)});
                            \draw [shift={(2.742622161145658,1.6668890701531305)},line width=.5pt]  plot[domain=-1.59153377431449:2.5486310581400553,variable=\t]({1.*1.8618051648137754*cos(\t r)+0.*1.8618051648137754*sin(\t r)},{0.*1.8618051648137754*cos(\t r)+1.*1.8618051648137754*sin(\t r)});
                            \draw [shift={(-2.465537155253595,-1.1298212608379852)},line width=.5pt]  plot[domain=1.5762493642788191:5.74230210194738,variable=\t]({1.*1.9645935077504308*cos(\t r)+0.*1.9645935077504308*sin(\t r)},{0.*1.9645935077504308*cos(\t r)+1.*1.9645935077504308*sin(\t r)});
                            
                            \draw [shift={(0.010019794693657956,0.24039996447748466)},line width=.5pt,dotted]  plot[domain=-0.8843461053335542:1.7802068292044266,variable=\t]({1.*1.1883729097106845*cos(\t r)+0.*1.1883729097106845*sin(\t r)},{0.*1.1883729097106845*cos(\t r)+1.*1.1883729097106845*sin(\t r)});
                            
                            \draw [shift={(-1.6721808953490571,-1.5359448703430314)},line width=.5pt]  plot[domain=1.5623326614435757:3.1948918559637867,variable=\t]({1.*2.132503383197742*cos(\t r)+0.*2.132503383197742*sin(\t r)},{0.*2.132503383197742*cos(\t r)+1.*2.132503383197742*sin(\t r)});
                            \draw [shift={(-1.7567191238010713,4.788179727741242)},line width=.5pt]  plot[domain=4.736857905199921:5.7305117057045125,variable=\t]({1.*4.1929527509820375*cos(\t r)+0.*4.1929527509820375*sin(\t r)},{0.*4.1929527509820375*cos(\t r)+1.*4.1929527509820375*sin(\t r)});
                            \draw [shift={(3.1197586444034977,1.8152012539590958)},line width=.5pt]  plot[domain=1.0165003155971395:2.608416189400072,variable=\t]({1.*1.5185318847315663*cos(\t r)+0.*1.5185318847315663*sin(\t r)},{0.*1.5185318847315663*cos(\t r)+1.*1.5185318847315663*sin(\t r)});
                        \end{tikzpicture}

                        \caption[A connected sum of spheres]{A connected sum of spheres. Cutting the manifold open along the solid line gives a space that is homeomorphic to the octagon that is shown in Figure~\ref{fig:connected_sum_of_2-manifolds}.}
                        \label{fig:visualising_a_connected_sum_of_spheres}
                    \end{figure}
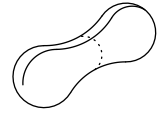

                    
                    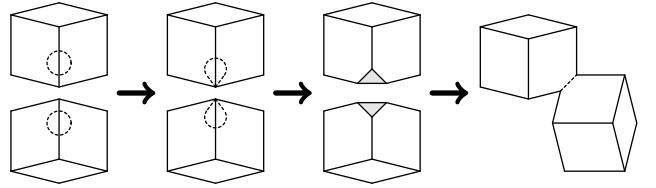
\begin{figure}[ht]
                        \centering

                        \begin{tikzpicture}[line cap=round,line join=round,>=triangle 45,x=1.0cm,y=1.0cm,scale=0.8]
                            \clip(-3.8,-1.8) rectangle (7.3,1.6);
                            \draw [line width=.5pt] (0.,0.)-- (0.8,0.2);
                            \draw [line width=.5pt] (0.,0.)-- (-0.8,0.2);
                            \draw [line width=.5pt] (0.,0.)-- (0.,1.);
                            \draw [line width=.5pt] (0.,1.)-- (0.8,1.2);
                            \draw [line width=.5pt] (0.8,1.2)-- (0.8,0.2);
                            \draw [line width=.5pt] (-0.8,0.2)-- (-0.8,1.2);
                            \draw [line width=.5pt] (-0.8,1.2)-- (0.,1.);
                            \draw [line width=.5pt] (-0.8,1.2)-- (0.,1.4);
                            \draw [line width=.5pt] (0.,1.4)-- (0.8,1.2);
                            \draw [line width=.5pt] (0.,-0.2)-- (0.,-1.2);
                            \draw [line width=.5pt] (0.,-0.2)-- (-0.8,-0.4);
                            \draw [line width=.5pt] (0.,-0.2)-- (0.8,-0.4);
                            \draw [line width=.5pt] (0.8,-0.4)-- (0.8,-1.4);
                            \draw [line width=.5pt] (0.8,-1.4)-- (0.,-1.2);
                            \draw [line width=.5pt] (0.,-1.2)-- (-0.8,-1.4);
                            \draw [line width=.5pt] (-0.8,-1.4)-- (-0.8,-0.4);
                            \draw [line width=.5pt] (-0.8,-1.4)-- (0.,-1.6);
                            \draw [line width=.5pt] (0.,-1.6)-- (0.8,-1.4);
                            \draw [line width=.5pt] (-1.8,0.2)-- (-2.6,0.);
                            \draw [line width=.5pt] (-2.6,0.)-- (-3.4,0.2);
                            \draw [line width=.5pt] (-2.6,0.)-- (-2.6,1.);
                            \draw [line width=.5pt] (-2.6,1.)-- (-1.8,1.2);
                            \draw [line width=.5pt] (-1.8,1.2)-- (-1.8,0.2);
                            \draw [line width=.5pt] (-3.4,0.2)-- (-3.4,1.2);
                            \draw [line width=.5pt] (-3.4,1.2)-- (-2.6,1.);
                            \draw [line width=.5pt] (-3.4,1.2)-- (-2.6,1.4);
                            \draw [line width=.5pt] (-2.6,1.4)-- (-1.8,1.2);
                            
                            
                            \draw [line width=.5pt] (3.4,0.2)-- (3.4,1.2);
                            \draw [line width=.5pt] (3.4,1.2)-- (2.6,1.);
                            
                            
                            \draw [line width=.5pt] (2.6,1.)-- (1.8,1.2);
                            \draw [line width=.5pt] (1.8,1.2)-- (1.8,0.2);
                            \draw [line width=.5pt] (1.8,1.2)-- (2.6,1.4);
                            \draw [line width=.5pt] (2.6,1.4)-- (3.4,1.2);
                            
                            
                            \draw [line width=.5pt] (2.36,0.06)-- (2.84,0.06);
                            \draw [line width=.5pt] (2.6,1.)-- (2.6,0.3);
                            \draw [line width=.5pt] (2.6,0.3)-- (2.36,0.06);
                            \draw [line width=.5pt] (2.6,0.3)-- (2.84,0.06);
                            \draw [line width=.5pt] (2.84,0.06)-- (3.4,0.2);
                            \draw [line width=.5pt] (2.36,0.06)-- (1.8,0.2);
                            
                            \draw [line width=.5pt,dash pattern=on 1pt off 1pt] (-2.6,0.4) circle (0.2cm);
                            \draw [line width=.5pt,dash pattern=on 1pt off 1pt] (0.,0.)-- (0.15,0.2);
                            \draw [line width=.5pt,dash pattern=on 1pt off 1pt] (0.,0.)-- (-0.15,0.2);
                            
                            \draw [shift={(0.,0.29982142857142857)},line width=.5pt,dash pattern=on 1pt off 1pt]  plot[domain=-0.5871779748174992:3.7287706284072923,variable=\t]({1.*0.1801785714285714*cos(\t r)+0.*0.1801785714285714*sin(\t r)},{0.*0.1801785714285714*cos(\t r)+1.*0.1801785714285714*sin(\t r)});
                            \draw [line width=.5pt,dash pattern=on 1pt off 1pt] (0.,-0.2)-- (0.15,-0.4);
                            \draw [line width=.5pt,dash pattern=on 1pt off 1pt] (0.,-0.2)-- (-0.15,-0.4);
                            \draw [shift={(0.,-0.4998214285714286)},line width=.5pt,dash pattern=on 1pt off 1pt]  plot[domain=-3.7287706284072923:0.5871779748174996,variable=\t]({1.*0.1801785714285714*cos(\t r)+0.*0.1801785714285714*sin(\t r)},{0.*0.1801785714285714*cos(\t r)+1.*0.1801785714285714*sin(\t r)});
                            
                            \draw [line width=.5pt] (-2.6,-0.2)-- (-3.4,-0.4);
                            \draw [line width=.5pt] (-2.6,-0.2)-- (-1.8,-0.4);
                            \draw [line width=.5pt] (-2.6,-0.2)-- (-2.6,-1.2);
                            \draw [line width=.5pt] (-1.8,-0.4)-- (-1.8,-1.4);
                            \draw [line width=.5pt] (-3.4,-0.4)-- (-3.4,-1.4);
                            \draw [line width=.5pt] (-3.4,-1.4)-- (-2.6,-1.6);
                            \draw [line width=.5pt] (-2.6,-1.6)-- (-1.8,-1.4);
                            \draw [line width=.5pt] (-1.8,-1.4)-- (-2.6,-1.2);
                            \draw [line width=.5pt] (-2.6,-1.2)-- (-3.4,-1.4);
                            
                            \draw [line width=.5pt,dash pattern=on 1pt off 1pt] (-2.6,-0.6) circle (0.2cm);
                            
                            
                            \draw [line width=.5pt] (2.6,-1.2)-- (3.4,-1.4);
                            \draw [line width=.5pt] (3.4,-1.4)-- (3.4,-0.4);
                            \draw [line width=.5pt] (2.6,-1.2)-- (1.8,-1.4);
                            \draw [line width=.5pt] (1.8,-1.4)-- (1.8,-0.4);
                            \draw [line width=.5pt] (1.8,-1.4)-- (2.6,-1.6);
                            \draw [line width=.5pt] (2.6,-1.6)-- (3.4,-1.4);
                            
                            
                            \draw [line width=.5pt] (2.36,-0.26)-- (2.84,-0.26);
                            \draw [line width=.5pt] (1.8,-0.4)-- (2.36,-0.26);
                            \draw [line width=.5pt] (2.36,-0.26)-- (2.6,-0.5);
                            \draw [line width=.5pt] (2.6,-0.5)-- (2.84,-0.26);
                            \draw [line width=.5pt] (2.84,-0.26)-- (3.4,-0.4);
                            \draw [line width=.5pt] (2.6,-0.5)-- (2.6,-1.2);
                            \draw [line width=.5pt] (4.4,0.)-- (4.4,1.);
                            \draw [line width=.5pt] (4.4,0.)-- (5.2,-0.2);
                            
                            
                            \draw [line width=.5pt] (6.,1.)-- (5.2,0.8);
                            \draw [line width=.5pt] (4.4,1.)-- (5.2,0.8);
                            \draw [line width=.5pt] (4.4,1.)-- (5.2,1.2);
                            \draw [line width=.5pt] (5.2,1.2)-- (6.,1.);
                            \draw [line width=.5pt] (5.2,0.8)-- (5.2,-0.2);
                            
                            
                            \draw [line width=.5pt] (6.,0.2)-- (6.8,0.2);
                            \draw [line width=.5pt] (5.7333333333333325,-0.06666666666666711)-- (5.6,-0.6);
                            \draw [line width=.5pt] (5.6,-0.6)-- (5.8,-1.4);
                            \draw [line width=.5pt] (6.8,0.2)-- (6.6,-0.6);
                            \draw [line width=.5pt] (6.6,-0.6)-- (6.8,-1.4);
                            \draw [line width=.5pt] (6.8,-1.4)-- (5.8,-1.4);
                            \draw [line width=.5pt] (5.6,-0.6)-- (6.6,-0.6);
                            \draw [line width=.5pt] (6.8,-1.4)-- (7.,-0.6);
                            \draw [line width=.5pt] (7.,-0.6)-- (6.8,0.2);
                            \draw [line width=.5pt] (5.2,-0.2)-- (5.7333333333333325,-0.06666666666666711);
                            \draw [line width=.5pt] (6.,0.2)-- (6.,1.);
                            
                            \draw [line width=.5pt,dash pattern=on 1pt off 1pt] (5.7333333333333325,-0.06666666666666711)-- (6.,0.2);

                            \fill[line width=.5pt,color=black,fill opacity=0.1] (2.36,0.06) -- (2.6,0.3) -- (2.84,0.06) -- cycle;
                            \fill[line width=.5pt,color=black,fill opacity=0.1] (2.36,-0.26) -- (2.84,-0.26) -- (2.6,-0.5) -- cycle;
                            
                            \draw [line width=2.pt,, -to] (-1.6,-0.1)-- (-1.,-0.1);
                            \draw [line width=2.pt,, -to] (1.,-0.1)-- (1.6,-0.1);
                            \draw [line width=2.pt,, -to] (3.6,-0.1)-- (4.2,-0.1);
                        
                        \end{tikzpicture}

                        \caption[Connected sum of 3-manifolds]{The connected sum of 3-manifolds is represented as the gluing of two cubes; this is a 3D generalisation of the 2D case that is depicted in Figure~\ref{fig:connected_sum_of_2-manifolds}. The dotted outlines represent the (spherical) boundaries of the open balls that have been removed. The triangular faces represent the cut boundaries after they have met the corner point of each cube, similarly as the cut circular boundary is encountered in Figure~\ref{fig:connected_sum_of_2-manifolds}; in this case, the corner point of the cube is the point to which the triangular boundary is mapped to.
                        }
                        \label{fig:connected_sum_of_3-manifolds}
                    \end{figure}
    
        \subsection{Multi-component samples}\label{subsec:multi-component_samples}
            Only single-component samples were used in
            ~\cite{Han23,Han23arXiv}. 
            These papers represented 4D cubes that contained cavities with 4D tensors of binary values. Future efforts could explore the use of multi-component samples ($\beta_0 > 1$) or samples that are not necessarily cubes with cavities. That is, rather than beginning with a tensor in which every entry set to~1 (a solid cube), each entry could begin as~0 in order to represent an empty space; this would offer a departure from the setting that was used in the referenced work. The space could then be filled with objects, including embeddings of
            3-manifolds and
            4-manifolds-with-boundary, by setting entries to 1. 
            Just as one could imagine embedding a circle in 3D space, one could
            potentially also embed circles, 2-spheres, tori, donuts, or other lower-dimensional manifolds (including those that may not be orientable) in the 4D space, since the idea of `cutting-out' would no longer be relevant.
            
            The approach that was proposed at the end of Section~\ref{subsubsec:computing_homology_groups_and_betti_numbers_of_connected_sums} to generate 4D tunnels (replacing an $S^1$ or $S^2$ factor with $I$ or $I^2$, respectively) could also be employed to describe 4D embeddings of tubes such as 
            $I \times S^2$, $I^2 \times S^1$, and $I \times S^1 \times S^1$. 
            3D embeddings of these tubes could also be considered and would appear as a thickened sphere, donut, and thickened torus, respectively, similarly as $I \times S^1$ can be embedded in a 2D space as an annulus and in 3D as a tube.

            \subsection{A roadmap towards 4D deformations}\label{subsec:roadmap_towards_4d_deformations}
                This section demonstrates two relatively basic approaches 
                to vary how features such as cavities and other manifolds could appear in the image-type data of 
                ~\cite{Han23,Han23arXiv}. 
                The first approach is a volume-preserving pixel-wise approach that takes inspiration from old techniques (namely, digital topology and pixel-wise deformations) and leverages later ideas (such as homology and persistent homology software) to identify suitable pixels.
                The second approach employs mathematical morphology.
                While the theoretical concepts that are required to implement these approaches in 3D and 4D are well-established, 
                the computational requirements of applying these approaches in 2D already suggest that 3D and 4D applications will be computationally demanding.
                
                \subsubsection{A hypervolume-preserving approach}\label{subsubsec:hypervolume-preserving_approach}
                    Hypervolumes (higher-dimensional volumes) were a key design element of the 
                    `4D mixed 128' 
                    dataset that was described in~\cite{Han23arXiv}; 
                    the hypervolume of a sample was taken as the number of 1-valued pixels.
                    One approach that could be used to vary the appearance of samples while preserving their hypervolume and homology is 
                    a pixel-moving algorithm, which when given a seed image containing various seed objects, could produce deformations that interact, for example, by swirling and wrapping objects around one another.
                    The idea is to select 1-valued boundary pixels of an object and displace them to suitable new 0-valued coordinates by flipping their values (setting 1 to 0 and vice versa),
                    provided that both removing the pixel and placing the pixel at its new location do not alter the homology of the neighbourhood of the flipped pixels; this check can be performed with persistent homology software and has a similar role to identifying simple pixels in digital topology.
                    Flipping pixels will preserve the sample's volume, and the chance of removing or introducing homological features is eliminated by preserving the homology of a pixel's neighbourhood.
                    In a 2D image, a neighbourhood would generally consist of a square of 9 pixels, with the selected pixel at its centre. 
                    
\begin{figure}[ht!]
                        \centering
                        
                        \includegraphics[scale = 0.4]{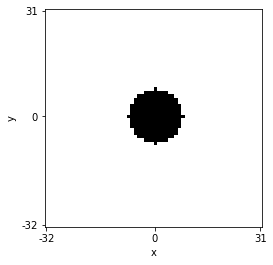}
                        \includegraphics[scale = 0.4]{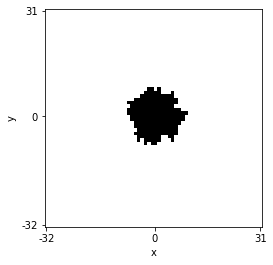}

                        \includegraphics[scale = 0.4]{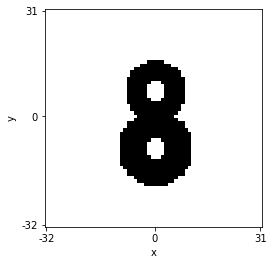}
                        \includegraphics[scale = 0.4]{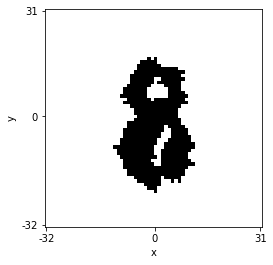}
                        
                        \caption[Deforming samples]{A disc (top left) and a figure-eight arrangement (bottom left) were used as seed objects and were deformed over 600 iterations to produce the results on the right.}
                        \label{fig:deformations_2d}
                    \end{figure}

                    A noise signal can be used to preferentially move and place pixels, for example, by moving 1-valued pixels that are at lower noise coordinates to higher noise 0-valued coordinates; this can produce the effect of thicker or bulging (conversely, thinner and narrowing) regions.
                    Other instructions could be implemented, to monitor thickness, curvature, or convexity~\cite{Tar22,Niy08}, or to dictate how far a pixel can be moved on each iteration.
                    Figure~\ref{fig:deformations_2d} shows the results of applying a 2D version of the hypervolume-preserving deformation approach to two $64^2$ seed images. An Improved Perlin noise~\cite{Per02} signal was generated for a space of the same dimensions so that each coordinate was attributed a noise value. On each iteration, a lower noise 1-valued boundary pixel was selected and moved to a higher noise 0-valued neighbour.
                    The resulting images show the effect after 600 iterations; 
                    this took approximately fifty minutes to complete on a 2021 MacBook Pro.
                    
                    Fitting a deformed object into a sample with appropriate spacing and without intersection can be achieved by employing morphological transformations~\cite{Har87,Dou92}, namely dilation.
                    Morphological transformations are image transformations that can be applied to grey-scale and binary images, and also include erosion, opening, closing, and gradient.
                    The dilation $\oplus$ of the object $M$ by a structuring element $B$ is found by adding each vector in $M$ to each vector in $B$, that is,
                    $ 
                        M \oplus B=\{ m+b \,|\, m \in M,\, b \in B \}.
                    $
                    An example of a structuring element is a ball with a radius that is equal to the chosen minimum spacing between objects.
                    Dilation may also be expressed as $M \oplus B = \bigcup_{b\in B} M_b$, where $M_b$ denotes the result of translating $M$ by the vector $b$; pilot experiments suggest that this interpretation appears to be the most straightforward to implement because iterating over the structuring element (which is presumably smaller than the seed image) usually requires less steps than iterating over a seed image.
                    The dilation results in a superimposition of $B$ at each 1-valued pixel in $M$, which, in effect, resembles $M$ with a buffer that is equal to the radius of $B$. The dilated object can be used during sample generation to find where in a sample is available to introduce $M$.

                \subsubsection{A morphology-based approach}
                    Two software strategies to achieve image skeletonisation are
                    decision-tree-based~\cite{McC14, Hom07, Lee94} and
                    kernel-based approaches, which employ morphological erosion (the dual of dilation) using a sequence of structuring elements.
                    Skeletons that are computed by means of the hit-or-miss transform~\cite[Chapter 4]{Dou92} can preserve an image's topology. The hit-or-miss transform utilises a pair of disjoint structuring elements, the first of which is used to perform erosion on the image itself and the second to erode the image with inverted pixels. The intersection of the resulting images is the output of the transform and the pixels that are present in the output image can be removed from the source image.
                    Depictions of how 3D and 4D structuring elements can appear are shown in~\cite{Jon00}.
    
\begin{figure}[ht!]
                        \centering
                        
                        \includegraphics[scale = 0.4]{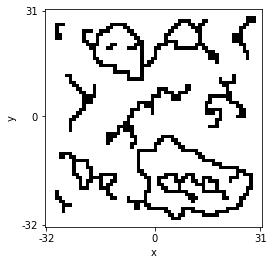}
                        \includegraphics[scale = 0.4]{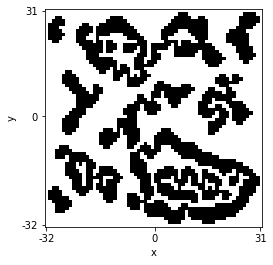}    
                        
                        \caption[Deforming a sample with morphology]{A skeletonisation approach was applied to a seed image (left) over 14 iterations to produce the image on the right.}
                        \label{fig:deformations_via_morphology}
                    \end{figure}

                    By inverting the task, skeletonisation can also be used to inscribe objects into a space. In 2D, for example, one can begin by introducing curves and circles into a seed image. The hit-or-miss transform can then be applied to the inverse of the image, that is, the transform will be applied to the background (0-valued) pixels, which thickens the seed curves and circles. Figure~\ref{fig:deformations_via_morphology} depicts a seed image comprising of an assortment of randomly drawn curves and circles, some of which have been glued together. The effect of the proposed method is shown in the image on the right; this took approximately one second to complete on a 2021 MacBook Pro. A 3D sample could be generated by randomly inscribing curves or small balls to produce 3-balls, and inscribing circles (that may be connected to one another or possess handles) to produce (multi-holed) donuts. 
                    Similarly, seed objects in 4D could consist of curves, circles, spheres, and tori to produce examples of $B^4$, $S^1 \times B^3$, $S^2 \times B^2$, and $S^1 \times S^1 \times B^2$, respectively. 
                    Inscribing a random curve or circle in a 4D seed image is straightforward and can be achieved similarly as done so in 2D. To inscribe deformed spheres and tori, one can begin in the 3D setting and leverage any suitable 3D approach. 

\begin{figure}[ht!]
                        \centering
                        
                        \includegraphics[scale = 0.4]{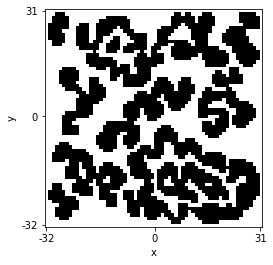}    
                        
                        \caption[Deforming a sample with morphology - alternate]{Three iterations of a modified morphology-based approach was applied to the seed image that is shown in Figure~\ref{fig:deformations_via_morphology}. }

                        \label{fig:deformations_via_morphology_balls}
                    \end{figure}

                    A similar effect to that which is shown in Figure~\ref{fig:deformations_via_morphology} can be produced by applying the standard dilation operation using a structuring element that represents a unit ball, with the modification that the operation is only performed if it does not alter the homology of the neighbourhood of the flipped pixels (similarly as described in Section~\ref{subsubsec:hypervolume-preserving_approach}). This option is potentially conceptually less complicated and simpler to implement than the hit-or-miss approach, particularly in dimensions higher than two, where the set of structuring elements that can be used for the hit-or-miss transform becomes larger~\cite{Jon00}. A 2D example that was produced by applying this modified morphology-based approach to the seed image that is shown in Figure~\ref{fig:deformations_via_morphology} with a structuring element that represents a unit 2-ball is presented in Figure~\ref{fig:deformations_via_morphology_balls}; this took approximately one minute to complete on a 2021 Apple MacBook Pro.

                    
\begin{figure}[ht!]
                        \centering

                        \includegraphics[trim=0 1cm 0 2cm, clip, scale = 0.25]{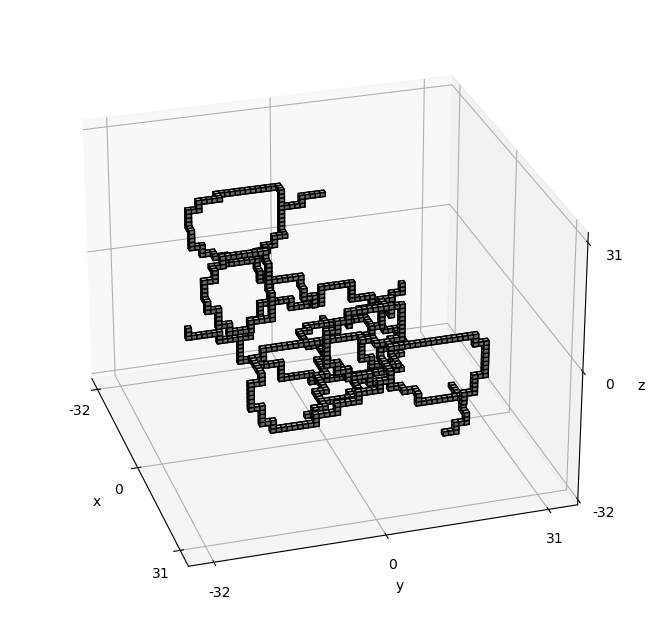}%
                        \includegraphics[trim=0 1cm 0 2cm, clip, scale = 0.25]{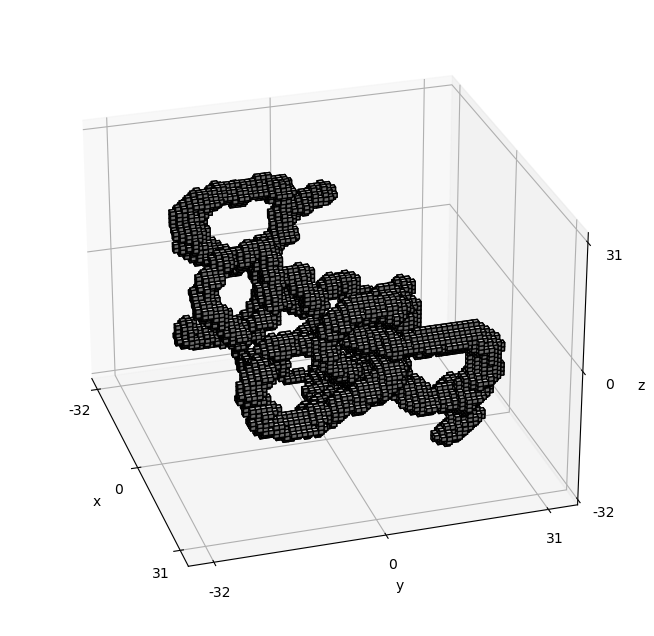}
                        
                        \includegraphics[trim=0 0.5cm 0 1.5cm, clip, scale = 0.25]{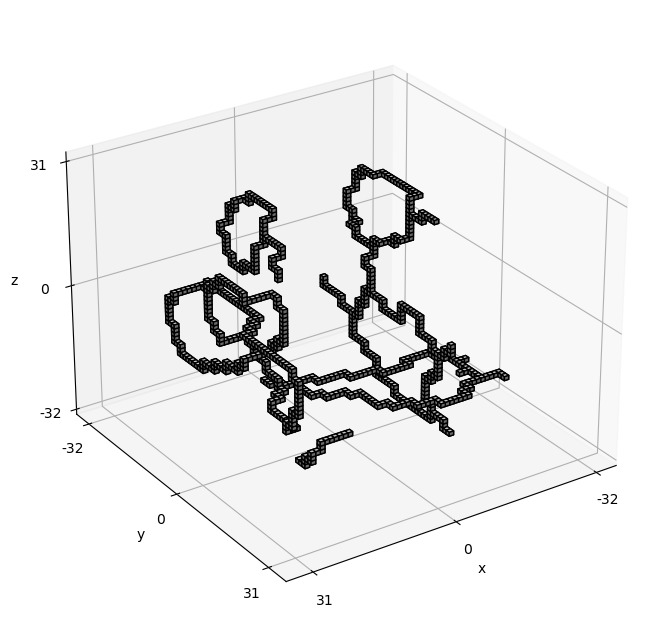}%
                        \includegraphics[trim=0 0.5cm 0 1.5cm, clip, scale = 0.25]{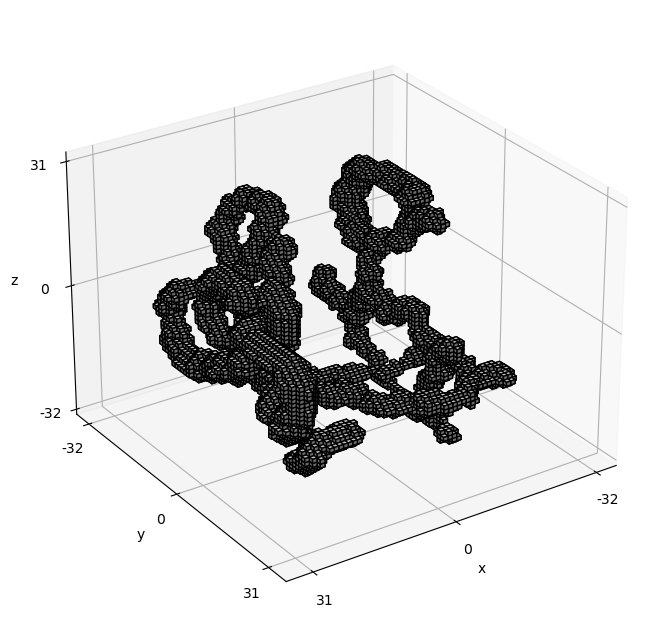}

                        \caption[Deforming a sample with morphology - 3D]{A 3D sample that is produced by applying 3 iterations of a modified morphology-based approach to a 3D seed image. The images on the left depict the same seed image from two viewpoints.
                        The seed image contains a curve (which can be seen at the front of the second row of images), several circles (two that have been connected can be seen in the centre of the second row), a link (seen in the bottom-right side of the images in the second row), and a (trefoil) knot; some of the objects have tentacles. These objects are reflected in the resulting image as a deformed ball and deformed donuts that are linked or knotted. The knotted donut is seen most clearly at the front of the first row. The Betti numbers for this image are $\beta_0 = 5$, $\beta_1 = 5$, and 0 otherwise.   }

                        \label{fig:deformations_via_morphology_3d}
                    \end{figure}

                    A 3D example is depicted in Figure~\ref{fig:deformations_via_morphology_3d}. 
                    The resulting 3D image took approximately five minutes to produce on the same hardware.
                    The same effect could be applied to a 4D seed image using a structuring element that represents a unit 4-ball. 
                    The modified morphology approach also makes it possible to diversify the data by not only using a variety of seed objects with interesting geometries, but by also varying how the dilation would be applied. For example, given a noise signal such as those in Section~\ref{subsubsec:hypervolume-preserving_approach}, the dilation could be randomly applied to a pixel to produce a lumpy result with thick and thin regions.

    \section{Conclusion}\label{sec:conclusion}
        There is interest and benefit in understanding the topological characteristics of data in many areas of research, 
        however, existing options that capture this information demonstrate practical limitations. The application of CNNs shows promise in this space.
        Applying topological labels to existing unlabelled data can be expensive and becomes challenging once data becomes too complex or extends past 3D. Synthetic data may prove to be a suitable alternative with which to train CNN models, particularly in scenarios where persistent homology may not be appropriate.
        This work offers an early look at how diverse and complex synthetic 4D image-type data with topological labels can be acquired. 
        Future research could explore how to efficiently implement the 4D ideas that were detailed in this work and apply them in dataset generation.

    \section*{Acknowledgements}
        This research was supported by the Australian Government through the Australian Research Council's Discovery Projects funding scheme 
        (DP210103304 `Estimating the Topology of Low-Dimensional Data Using Deep Neural Networks'). 
        The views expressed herein are those of the authors and are not necessarily those of the Australian Government or Australian Research Council. The first author was supported by a PhD scholarship from the University of Newcastle associated with this grant.

    \bibliographystyle{IEEEtran}
    \bibliography{IEEEabrv, 4D}

\end{document}